# ColdGAN: Resolving Cold Start User Recommendation by using Generative Adversarial Networks

Po-Lin Lai, Chih-Yun Chen, Liang-Wei Lo, Chien-Chin Chen
Department of Information Management
National Taiwan University, Taipei Taiwan
{R08725022, R08725031, R08725024, patonchen}@ntu.edu.tw

## Abstract

Mitigating the new user cold-start problem has been critical in the recommendation system for online service providers to influence user experience in decision making which can ultimately affect the intention of users to use a particular service. Previous studies leveraged various side information from users and items; however, it may be impractical due to privacy concerns. In this paper, we present ColdGAN, an end-to-end GAN based model with no use of side information to resolve this problem. The main idea of the proposed model is to train a network that learns the rating distributions of experienced users given their cold-start distributions. We further design a time-based function to restore the preferences of users to cold-start states. With extensive experiments on two real-world datasets, the results show that our proposed method achieves significantly improved performance compared with the state-of-the-art recommenders.

## 1 INTRODUCTION

With the rapid growth of internet services such as e-commerce, streaming platforms, and social networks, recommendation systems(RS) have been indispensable for people because of the ability to mitigate the difficulties of information overload effectively [16,17,21]. A great deal of surveys and studies (e.g., [2,6,7]) prove that personalized internet services via item recommendations are able to not only improve user experience but also gain customer loyalty that further increases service revenues. Consequently, studying recommendation systems is practical and important.

Recently, Generative Adversarial Networks (GAN) [3], an advanced deep learning methodology, have gained a great deal of attention of recommendation system researchers due to its superior recommendation performance [8, 9, 15]. The principle of GAN is to play a minimax game between a *generator* and a *discriminator* that the generator strives to fool the discriminator by generating plausible samples similar to real data. In terms of recommendation systems, Tong et al. [8] developed a collaborative generative adversarial network (CGAN) to reduce the adverse impact of data noise and achieve improved performance. In this work, they use the rating vector for each item as input and incorporate Variational Auto-Encoder (VAE) [23] as a generator. After encoding, the generator learns the rating distribution from the training data and generates latent factors containing user-item interactions, then decoding the generated latent factors as the reconstructed rating vector employed to recommend items to the user. CFGAN [15] was the first model using vector-wise training with GAN in RS for generating a fake purchase vector (implicit feedback) to deceive the discriminator. They adopted a masking mechanism to address the sparsity by dropping out non-purchased items and a negative sampling technique from unobserved to avoid the generator find an easy solution that is outputting 1 for all elements in the purchase vector. Chae et al. [9] presented RAGAN, a GAN-based rating augmentation model, which utilizes the observed user preferences (e.g., ratings of items) to initiate the model parameters and generates plausible ratings by the learned generator. Next, an augmented matrix containing the observed and generated ratings is established to construct a Collaborative Filtering(CF) model for item recommendations. While the GAN-based recommendation methods achieve remarkable results, they may not be good at resolving the new user cold-start recommendation problem [4]. In these works, the user representations are constructed only depend on their historical interactions with items. Compared to warm users, recommendation systems generally have sparse rating history of new users thus they are unable to make satisfying recommendations for a new user.

Several methods have been developed to address the new user cold-start problem. Most of the methods resolve the rating sparsity of cold-start users by leveraging various side information, such as user demographic data (e.g., age, gender, location etc.) [10,12], item features (e.g., text description, audio or image) [10,12], and social relationships [11] to generate proper recommendations for new users. For instance, Volkovs et al. [10] aim to build a model that can deal with both warm and cold user scenarios. For each user, they utilized user preferences (e.g., ratings) and content information as inputs for deep neural networks(DNNs), then concatenated them together and passed through another fine-tuning network to obtain user latent representation. They applied input dropout by setting rating inputs 0 at a ratio during network training to simulate cold-start situations, thus the model would learn to utilize content information to reproduce the accuracy of latent representation with available rating preference. Liang et al. [12] designed a Jointed Training Capsule Network(JTCN) to capture preference from side information for dealing with complete cold-start users problem. They extract user rating preference from embeddings of historical descriptions of items and obtain user content from user documents to form user representation, then jointly optimizing the mimic loss for generating user rating preference and the softmax loss for item recommendations. When predicting for a new user, the model can generate a comprehensive representation for the user based on the side information. Sedhain et al. [11] proposed a neighborhood-based CF method that directly exploits social networks of the user such as Facebook friends and page likes to compute user-user similarity.

While the experiments of the studies demonstrate that side information are useful to discover user preferences so as to recommend appropriate items to cold-start users, privacy concerns would make them impractical. Xu et al. [13] considered the rating made by cold-start users are more valuable than those of warm users. Their idea was first instantiating with Matrix Factorization(MF) [14] to obtain latent profiles of warm users and latent profiles for all items. With latent profiles of item fixed, updating latent profiles of cold-users by comparing the rating score between warm users and cold-start user. The latent profiles then are used to recommend items to cold-start users. Even though the method conducts cold-start recommendation without using side information, they are not suitable for capturing sequential patterns from past preferences of users because MF does not model the order of actions [24]. Besides, using the dot product as a similarity measure may not be sufficient to capture the complex structure of user interaction data.

In this paper, we resolve the new user cold-start recommendation problem by presenting a GAN-based method that makes no use of side information. Different to RAGAN that the employed GAN model is independent to the CF recommendation process, the proposed method is an end-to-end GAN approach. We only examine the ratings made by users to train a GAN model in which the generative network mimics (generates) the rating distributions of warm users given their cold-start distributions, and the discriminative network acts as a detector to distinguish the generated ratings from the real ratings. Also, a time-based rejuvenation function is designed to restore users' item ratings back to their cold-start states, and is incorporated into the GAN model for efficient model training. The learned generative network functions as a recommender to suggest items useful to new users. While the method in [10] have employed the drop-out mechanism that randomly drops item ratings to form cold-start states of users, our rejuvenation function further considers the temporal information of ratings that help retain crucial ratings for model learning. The experiment results based on the MovieLens datasets and Amazon Gift Card datasets show that the proposed GAN-based recommendation method is superior to state-of-the-art cold-start recommendation systems.

## 2 THE PROPOSED MODEL

Figure 1 shows the proposed new user cold start recommendation method which consists of two major components: *the GAN training* and *the cold-start recommendation generation*. We consider a user a *cold-start user* if the user has little item ratings. Conversely, a *warm user* is with many item ratings. In the GAN training phase, we collect a set of warm users along with their item ratings. Basically, warm users were cold-start when they just entered the recommendation system. They evolve into warm users as their item ratings accumulated. We design a time-based rejuvenation equation that drops the ratings made by a warm user. The earlier the rating made the less chance the rating is dropped to restore the warm user back to the cold-start state. Next, the ratings of the warm users together with the rejuvenated cold-start states are fed into a GAN model to learn a generative network *G* and a discriminative network *D*. The network *G* strives to produce plausible warm states given the cold-start states while the discriminative network discriminates the real warm states from the plausible states. After the generative network is learned, the cold-start recommendation generation uses it to suggest items relevant to the interests of a new user. Moreover, in order to enhance the recommendation result, we revise the loss function of the GAN model. We detail our method in the following sections.

# ColdGAN

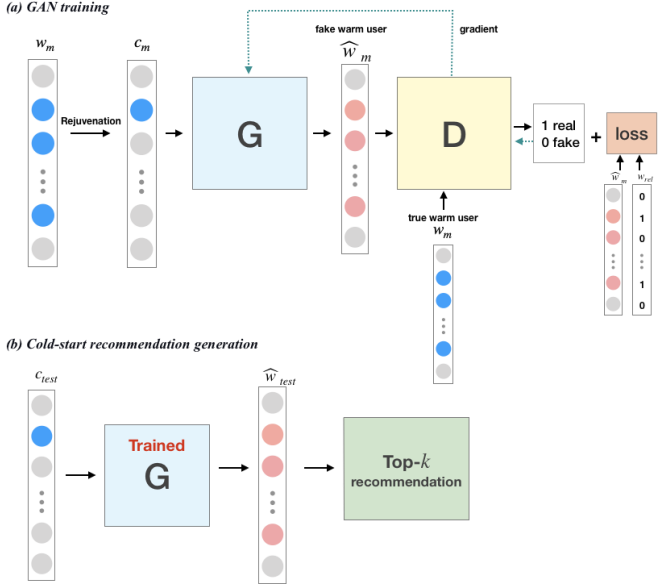

Figure 1: The architecture of our proposed model. (a) is the training phase of our proposed model where generative network $G$ infer warm states from cold-states. (b) is the testing phase to recommend relevant items for cold-start user.

## 2.1 The GAN Training

GAN is an advanced deep learning technique that establishes a minimax two-player game for training two adversarial networks, namely, the generative network $G$ and the discriminative network $D$. Given imperfect data (e.g., images corrupted with noise), the generative network learns to re-produce perfect data (e.g., original images) from the imperfect data and the generated data are evaluated by the discriminative network trained for detecting data not perfect enough. GAN has a huge impact on deep learning because it enables supervised learning to be unsupervised by generating perfect data automatically [3,25,26]. The merit of perfect data generation also motivates us in using GAN to resolve the new user cold-start recommendation problem. In this study, we consider the states (i.e., the item ratings) of warm users as perfect data. Our rejuvenation function corrupts a warm state by turning it back to the cold-start state. The proposed GAN model learns the rating distributions of the cold/warm states so the learned generative network would be able to predict items relevant to the interests of a new user.

Specifically, let $U = \{u_1, \ldots, u_M\}$ be a set of warm users and $I = \{i_1, \ldots, i_N\}$ be the items of the recommendation system. Like [19,20,22], we represent the warm state of a user $u_m$ as a *rating vector* $w_m \in \mathbb{R}^n$ that records the ratings of items made by $u_m$. Next, we rejuvenate the warm state $w_m$ to a cold-start state $c_m$ according to the following rejuvenation function:

$$p_m(i) = p_{min} + (p_{max} - p_{min}) * e^{\left(-\alpha * \frac{i}{count(w_m)}\right)} \quad (1)$$

where $p_m(i)$ indicates the probability that the $i$-th item rated by $u_m$ will be retained in $c_m$. Function $count$ returns the number of rated items in $w_m$. Symbols $p_{min}, p_{max}$ and $\alpha$ are real numbers and $0 \leq p_{min} < p_{max} \leq 1$. Figure 2 shows the distribution of our time-based rejuvenation function that the earlier the items were rated, the more likely it will be retained in $c_m$ and vice versa. The warm states of the users in $U$ and the corresponding cold-start states are then fed into the GAN model to simulate the process of how a cold-start user becomes a warm user.

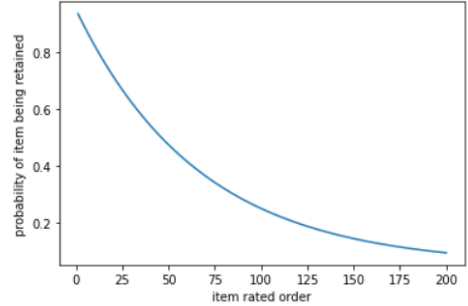

Figure 2: The distribution of our time-based rejuvenation function.

Our GAN model respectively employs the denoising autoencoder (DAE) [18] and the multi-layer perceptron (MLP) to construct the generative network and the discriminative network. In the model training, we first randomly initiate the weights of the networks (i.e., the model parameters) and then iterate the following two steps until the training loss defined in Equation (2), (3) is converged.

i. Training network $D$: In this step, we fix the weights in $G$ and regard $D$ as a binary classifier. The input of $D$ is a set of $\hat{w}_m$ and $w_m$ that the network is learned to classify true warm states.

ii. Training network $G$: The input of $G$ is a set of $c_m$'s that is flourished by $G$ to output a plausible warm state $\hat{w}_m$ for each $c_m$. In this step, the weights in $D$ are fixed, and the training process adjusts $G$'s weights to maximizing $D(\hat{w}_m)$, that is, to enable $G$ deceiving $D$.

The loss functions of our $G$ and $D$ networks are as follows:

$$\mathcal{L}_{GAN}^D = -E_{x \sim P_{warm}}[\log(D(x)) - E_{\hat{x} \sim P_g}[\log(1 - D(\hat{x}))] \quad (2)$$

$$\mathcal{L}_{GAN}^G = -E_{\hat{x} \sim P_g}[D(\hat{x})] + \mathcal{L}_{relevant} \quad (3)$$

where

$$\mathcal{L}_{relevant} = \sum_{i=1}^{m} \ell(\sigma(w_m), w^{rel})$$

where $D(x)$ in Eq. (2) and (3) indicates the estimated probability of $x$ being in the ground truth (i.e. warm states), we optimize the $D$ networks to identify generated fake warm users $\hat{w}_m$ and true warm user $w_m$. $\mathcal{L}_{GAN}^G$ in Eq. (3) contains

ColdGAN .

two loss function. The former is the original loss of G optimize the probability to generate fake warm users $\hat{w}_m$ such that fixed D cannot tell the difference of $\hat{w}_m$ and $w_m$, the latter is our revised loss $\mathcal{L}_{relevant}$ used to mitigate the limitation of common problems in GAN. We will discuss it in next section.

In summary, we feed the cold-state (i.e., $c_m$) of each warm user into G to generate a fake warm user $\hat{w}_m$, which is expected to contain ratings that $u_m$ would like to give in the future. D aims to identify generated fake warm users $\hat{w}_m$ and true warm user $w_m$.

**Relevant items Loss**
Although vanilla GAN has shown the ability to approximate real data distribution, there are several shortcomings as vanishing gradients and mode collapse that lead it hard to train [27,28]. Mode collapse happens when the generative network only learns to generate warm rating vectors from a few outputs of the rating distribution of warm states, but misses many others, even though missed outputs exist throughout the training data. More precisely, while G generates a fake warm user to deceive D, it may tend to generate a plausible vector but not relate to the specific user. Therefore, we modify our GAN-based model by adding a relevant item loss $\mathcal{L}_{relevant}$ to jointly optimize our model. We denote a n-dim binary vector $w^{rel}$ where an element is 1 if the item has been rated and relevant to the warm user and 0 otherwise. $\ell$ is Binary Cross-Entropy loss as the criterion for $\sigma(\hat{w}_m)$ and $w^{rel}$.

This loss aims to let the generative networks G not only learns to generate plausible vector, but also focus the relevant items for specific user.

**2.2 The Cold-Start Recommendation Generation**
The generative network G is trained to infer warm states from cold-start states. Hence, after the GAN training, we utilize the generative network G to recommend items for cold-start new users. As shown in Figure 1 (b), for a testing cold-start user, we again represent the cold-start state as a rating vector $c_{test}$. The vector is fed into G to produce $\hat{w}_{test}$ that contains the ratings the cold-start user would make in the future. Then, except the items that have been rated in $c_{test}$, we rank all the items according to their ratings in $\hat{w}_{test}$ and recommend top-N items to the user.

# 3 EXPERIMENTS
## 3.1 Experimental Settings

| Datasets | # user | # item | # rating | scale | Sparsity |
|---|---|---|---|---|---|
| ML1M | 6,040 | 3,706 | 1,000,209 | [1,2,3,4,5] | 95.5% |
| Amazon | 198 | 3284 | 11,470 | [1,2,3,4,5] | 98.2% |

**Table 1. Statistics of datasets**

We conduct experiments on two real-world datasets: MovieLens(1M) and Amazon (Gift cards). For Amazon dataset, we filter the users with rating interactions less than 15 and items rated less than 3 times and finally get 198 users, 3,284 items and 11,470 interactions. Their details are shown in Table 1.

For each dataset, we randomly split 80% into train data and rest 20% for test data. In the testing phase, we select only 10 earliest rated ratings (based on item timestamp) for each user and discarded the rest of her ratings in both datasets as the inputs. Besides, to evaluate effectiveness of top-$k$ recommendation, we adopt evaluation metrics: precision(P@k), recall (R@k), and normalized discounted cumulative gain(nDCG@k). We set k as 5,10.

Furthermore, we consider an item relevant (ground truth) if its rating score is greater than the mean value from rated item of the user.

## 3.2 Results and Analyses
### 3.2.1 *Model Comparisons*
Our proposed model compares with several representative recommendation models as below:

- **MF [14]** embeds user and item ID as vectors and measure user-item interaction with the inner product. We train the latent factor model with Binary Cross Entropy loss.
- **RAPARE-MF [13]** employs the ratings from existing users (warm users) to help calibrate the latent profiles of cold-start users. After updating latent profiles of cold users, apply MF to measure user-item interaction.
- **DAE [18]/CDAE [22]** both apply the Denoising Auto-encoder to item recommendation, which adds random noise to the input data before feeding into the auto-encoder, can be divided into with the user-specific vectors (CDAE) and without the user-specific vectors(DAE).
- **CGAN [8]** is GAN-based, which incorporates VAE as a generator. The decoder in generator will reconstruct the generated latent factors into rating vector of the user then use it to recommend items.
- **CFGAN [15]** adopts the idea of Conditional GAN. Inputs are implicit feedback. During the training, randomly sample negative items from unobserved to balance training data.
- **RAGAN [9]** is a GAN-based rating augmentation model. Augmented the rating matrix with generated plausible ratings by GAN and then use the matrix to train a MF model.

We follow the settings of the paper to build models mentioned above and tune the hyper-parameter to achieve best performance.

ColdGAN

| Datasets | ML1M | | | Amazon | | |
|---|---|---|---|---|---|---|
| metrics | p@5 | r@5 | ndcg@5 | p@5 | r@5 | ndcg@5 |
| MF | 0.0522 | 0.0038 | 0.0087 | 0.0050 | 0.0006 | 0.0000 |
| RAPARE | 0.0725 | 0.0046 | 0.0013 | 0.0250 | 0.0020 | 0.0000 |
| DAE/CDAE | 0.2432/ 0.2404 | 0.0202/ 0.0264 | 0.0783/ 0.0610 | 0.0688/ 0.0625 | 0.0120/ 0.0104 | 0.0247/ 0.0260 |
| CGAN | 0.2973 | 0.0280 | 0.0800 | 0.0750 | 0.0230 | **0.0500** |
| CFGAN | 0.1750 | 0.0082 | 0.0234 | 0.0281 | 0.0036 | 0.0028 |
| RAGAN | 0.1582 | 0.0121 | 0.0649 | 0.0100 | 0.0001 | 0.0020 |
| Our Model | **0.3625** | **0.0430** | **0.1124** | **0.1125** | **0.0307** | **0.0500** |

**Table 2. Comparisons with methods.**

Table 2 report the experimental results of model comparison on ML1M and Amazon dataset in terms of precision, recall and ndcg@5. The other metrics exhibited very similar tendency, hence they are omitted due to space limitations. Overall, we see that our proposed model significantly outperforms all other compared model on both datasets.

It's easy to observe that RAPARE, which is dedicated to cold-start problems and trained without any side information, performs slightly better than original MF. MF shows poor performance in the cold-start scenario since the user latent of cold-start users haven't been learned. Moreover, we observe that compared models with MF-based for item recommendations such as MF, RAPARE, and RAGAN. It may because the NN-based model has better representation learning capability due to its non-linear nature.

Besides, all other deep-learning based models perform better than RAPARE, which indicates the advantage of discovering patterns of sequential data in the NN-based model.

The relative performance improvements over CDAE(DAE), indicates that our proposed GAN is an effective extension to CDAE. Because we have shared a similar idea of the drop-out mechanism to construct a more robust user representation. This also demonstrates the adversarial training boosts the recommendation models in terms of accuracy. In particular, compared with other models that also adopt GAN technique, our proposed model consistently outperforms. In contrast, we leverage GAN and time-based rejuvenation to generate more plausible and personalize ratings for cold-start users.

3.2.2 *Effectiveness of rejuvenation and relevant loss*

This subsection investigates the impact of the following component which are unique in our proposed model: (1) the drop-out mechanism and (2) relevant item loss.

| Datasets | ML1M | | | Amazon | | |
|---|---|---|---|---|---|---|
| metrics | p@5 | r@5 | ndcg@5 | p@5 | r@5 | ndcg@5 |
| random | 0.3470 | 0.0413 | 0.0986 | **0.1250** | **0.0323** | **0.0542** |
| time-based | **0.3625** | **0.0430** | **0.1124** | 0.1125 | 0.0307 | 0.0500 |

**Table 3. Comparisons with different drop-out mechanism.**

Table 3 reports the experimental results on different drop-out mechanism. We observe that the proposed model works better in ML1M dataset, and achieve similar performance in Amazon dataset. It may because of the ability of the time-based distribution to retain more informative items for the generative network to learn what items the cold-user would like to rate in the future.

| Datasets | ML1M | | | Amazon | | |
|---|---|---|---|---|---|---|
| metrics | p@5 | r@5 | ndcg@5 | p@5 | r@5 | ndcg@5 |
| w/o | 0.3139 | 0.0295 | 0.1184 | 0.0313 | 0.0122 | 0.0111 |
| w | **0.3625** | **0.0430** | **0.1124** | **0.1125** | **0.0307** | **0.0500** |

**Table 4. Comparisons with/without relevant item loss**

Table 4 shows the performance with/without additional relevant item loss. This result shows relevant item loss could reach higher accuracy, especially on the sparse dataset(Amazon). It demonstrates that our approach effectively deals with shortcomings in vanilla GAN, our generative network not only learn to generate plausible rating but also focus on the specific user preference.

4 **CONCLUSIONS**

In this paper, a novel end-to-end GAN-based model, ColdGAN, is first introduced to infer experienced user ratings from their cold-start states with no use of side information for addressing new user cold-start problems. The proposed time-based rejuvenation function restores users' item ratings to their cold-start states and incorporates them into the GAN model for efficient model training. Moreover, we further revise the loss function of the original GAN to mitigate the Mode collapse problem. Extensive experiments on two real-world datasets not only demonstrate the enhanced accuracy but the effectiveness of rejuvenation and relevant loss, which significantly outperforms the state-of-the-art approaches in the cold-start recommendation system. Notably, the proposed model further enables future work on applying GAN to the new item cold-start problems.

ColdGAN